%% file: main.tex
\pdfoutput=1
\documentclass[11pt]{article}
\usepackage[]{acl}

\usepackage{times}
\usepackage{latexsym}
\usepackage{multirow}
\usepackage{booktabs}
\usepackage{soul}
\usepackage{hyperref}
\usepackage[inkscapeformat=png]{svg}
\usepackage[T1]{fontenc}
\usepackage{todonotes}

\usepackage[utf8]{inputenc}
\usepackage{microtype}
\usepackage{graphicx}
\usepackage{amsmath}
\usepackage{amssymb}
\usepackage{arydshln}

\usepackage[utf8]{inputenc}

\definecolor{ForestGreen}{RGB}{34,139,34}

\usepackage{array} 
\usepackage{colortbl} 
\usepackage{pdflscape} 

\newcolumntype{Y}{>{\centering\arraybackslash}X}

\usepackage{booktabs}
\usepackage{multirow}
\usepackage{graphicx}
\usepackage{subcaption}

\title{Intent-conditioned and Non-toxic Counterspeech Generation using Multi-Task Instruction Tuning with RLAIF}

\author{Amey Hengle$^{1}$\thanks{* Equal contribution}, Aswini Kumar$^{1*}$, Sahajpreet Singh$^{1}$, \textbf{Anil Bandhakavi}$^3$,  \\ \textbf{Md Shad Akhtar}$^2$, \textbf{Tanmoy Chakroborty}$^1$\\
$^1${IIT Delhi, India},
$^2${IIIT Delhi, India},
$^3${Logically.ai} \\
\small {
\{
\texttt{ameyhengle22},
\texttt{aswinikumarpadhi1995}, 
\texttt{sahaj.phy}
\}
\texttt{@gmail.com}
}\\
\small
\texttt{shad.akhtar@iiitd.ac.in},
\texttt{anil@logically.ai}, {\tt tanchak@iitd.ac.in}
}

\begin{document}
\maketitle

\input{sections/abstract}
\input{figures/figure1}

\input{sections/introduction}
\input{sections/related_work}
\input{tables/table1}
\input{sections/dataset}
\input{figures/figure2}
\input{sections/proposed_methodology}
\input{tables/results_table}
\input{sections/experimental_setup}
\input{sections/results_and_discussion}
\input{tables/human_eval}
\input{sections/conclusion}
\input{sections/limitation}
\input{sections/ethics_statement}

\bibliography{anthology}

\appendix
\input{sections/appendix}
\end{document}

%% file: sections/abstract.tex
\begin{abstract}
Counterspeech, defined as a response to mitigate online hate speech, is increasingly used as a non-censorial solution. Addressing hate speech effectively involves dispelling the stereotypes, prejudices, and biases often subtly implied in brief, single-sentence statements or abuses. These implicit expressions challenge language models, especially in seq2seq tasks, as model performance typically excels with longer contexts. 
Our study introduces \texttt{CoARL}, a novel framework enhancing counterspeech generation by modeling the pragmatic implications underlying social biases in hateful statements. \texttt{CoARL}'s first two phases involve sequential multi-instruction tuning, teaching the model to understand intents, reactions, and harms of offensive statements, and then learning task-specific low-rank adapter weights for generating intent-conditioned counterspeech. The final phase uses reinforcement learning to fine-tune outputs for effectiveness and non-toxicity. \texttt{CoARL} outperforms existing benchmarks in intent-conditioned counterspeech generation, showing an average improvement of $\sim$$3$ points in intent-conformity and $\sim$$4$ points in argument-quality metrics. Extensive human evaluation supports \texttt{CoARL}'s efficacy in generating superior and more context-appropriate responses compared to existing systems, including prominent LLMs like ChatGPT.
\end{abstract}

%% file: figures/figure1.tex
\begin{figure}[t]
\includegraphics[width=\columnwidth]{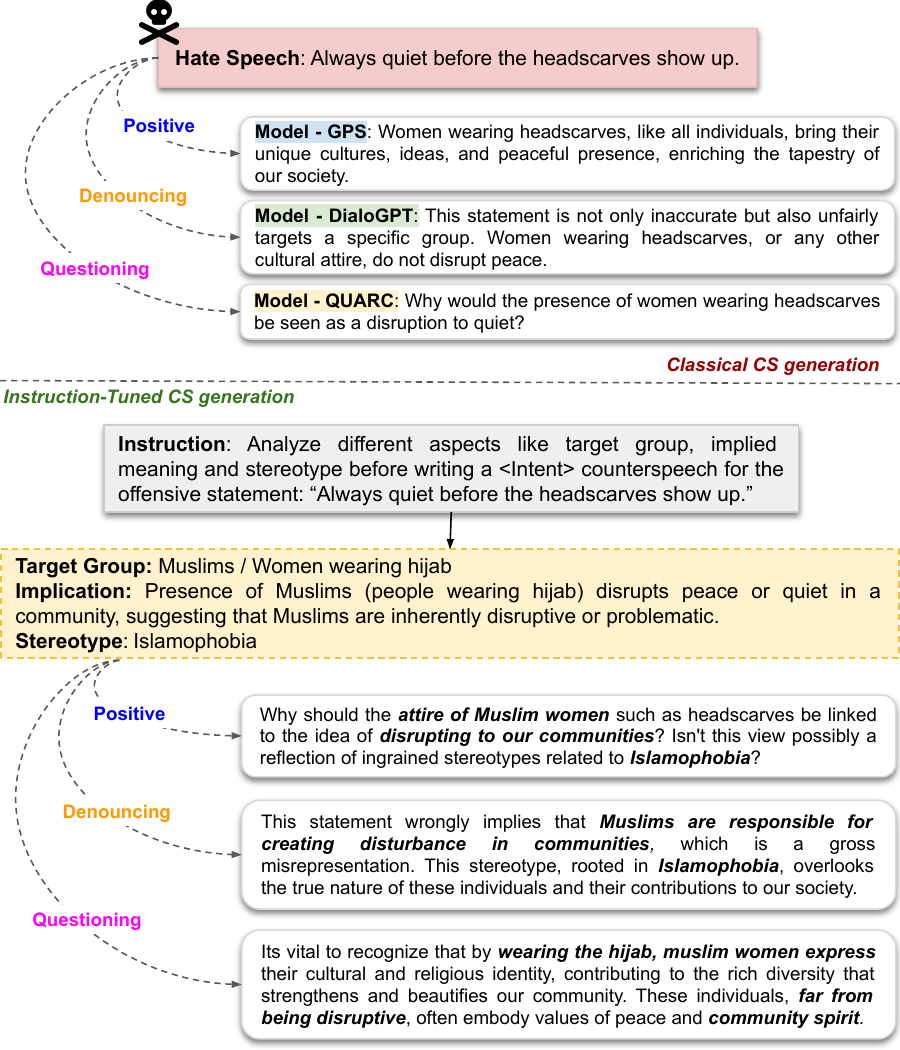}
\caption{Classical methods vs. instruction tuning for counterspeech generation. These examples show that counterspeech generation can be improved by the use of detailed and explicit instructions that allow a  model to focus on the different aspects of a given hate speech.}
\label{fig:figure1}
\vspace{-5mm}
\end{figure}

%% file: sections/introduction.tex
\section{Introduction}

Counterspeech (\texttt{CS}), defined as responses that counteract hate speech by seeking to undermine, weaken, or rebut hateful or offensive speech through the use of positive or constructive dialogue \cite{benesch2016-considerations-for-succesful, chandrasekharan2017you}, has proven to be an effective method for mitigating online hate while maintaining a diversity of voices and opinions \cite{schieb2016governing, wright2017vectors,masud2024probing}. This approach emerges as a more viable solution to address hateful online speech, avoiding the censorship risks associated with deletion-based content moderation. However, given the increasing scale of hateful content online \cite{leetaru-2019, masud2021hate,masud2022proactively,kulkarni2023revisiting}, relying solely on human-generated counterspeech is becoming increasingly tedious. In this scenario, NLP systems offer a promising avenue for understanding and automating the generation of counterspeech. Such systems could significantly aid content moderators and other stakeholders in efficiently and effectively countering online hate \cite{parker-et-al-2023,garg2023handling}. Consequently, there has been an increasing interest in research focusing on the detection, analysis, and generation of counterspeech \cite{mathew2019thou, qian2019benchmark, chung-et-al-2023-understanding-counterspeech, fanton2021human, bonaldi-etal-2022-human}. 

Generative approaches predominantly model it as a seq2seq problem, mirroring the structure of hate speech and its countering responses \cite{chung2019conan, sheng2020towards, zhu-bhat-2021-generate}. This approach, however, has evolved with the recognition that hate speech, and thus its counterspeech, is not monolithic. Different instances of hate speech may necessitate distinct types of counterspeech, tailored to the specific context and nature of the hateful content \cite{benesch2016-considerations-for-succesful,chung-et-al-2023-understanding-counterspeech}. This has inspired generative approaches that incorporate stylistic strategies (e.g., politeness, joyfulness, detoxification \cite{saha2022countergedi} and relevance \cite{sheng2020towards}). In particular, \citet{gupta-etal-2023-counterspeeches}  explored the concept of \textit{intent-specific} counterspeech generation, where the generation is conditioned on certain well-established counterspeech strategies. This approach offers a more nuanced and effective toolkit for moderators, providing them with a range of response options to counter hate \cite{benesch-2016-vectors-for-counterspeech}.

\paragraph{\large Motivation:}
As outlined by \citet{benesch2016-considerations-for-succesful}, effective counterspeech should not only align with a specific intent or strategy but also dispel any bias, prejudice, or stereotypical beliefs expressed in the hate speech. However, a substantial portion of online hate speech is characterized by brief, single-sentence statements or abuses. Furthermore, biases or stereotypical beliefs are rarely projected in what is stated explicitly, but rather through layers of implied meanings, subtly framing and influencing social judgments about certain groups \cite{social-bias-frames}. These short, often implied expressions of hate, pose a unique challenge to language models, which typically perform better with more extended contexts. This limitation is particularly pertinent in seq2seq modeling tasks, where the brevity of input can adversely affect the quality of counterspeech generation \cite{keneshloo2019deep-short-context}. 

Instruction tuning (IT) has been shown to improve over traditional supervised fine-tuning by providing explicit and detailed instructions to the language model \cite{zhou2023-InstructCTG}.  By providing explicit instructions, IT can help reduce the ambiguity in the input, making it easier for the model to understand the user's intent and generate more accurate outputs. We argue that counterspeech generation can be improved by adopting a similar setup. By providing clear and specific instructions on how to generate a desired counterspeech, IT can aid a language model to understand both the context and implied nuances of hate speech more effectively and, thus, can lead to more accurate and relevant responses. 

We support our argument by providing an example in Figure \ref{fig:figure1}, where for a given hate speech, we contrast the responses of three popular counterspeech generation models -- Generate-Prune-Select (GPS)  \cite{zhu-bhat-2021-generate}, DialoGPT \cite{zhang2020dialogpt}, and QUARC \cite{gupta-etal-2023-counterspeeches} against our proposed method, which employs IT instead of the classical supervised fine-tuning setup. We observe that although responses generated by classical methods are semantically coherent and somewhat aligned to the desired intent, they are either generic or fail to form convincing arguments against hate speech. This suggests the inability of the classical methods to capture certain implied aspects like bias, stereotype, or target group from such short statements. On the other hand, we observe in our IT setup that clear and explicit instructions help the model understand which aspects of hate speech to focus on and what type of CS is expected, which reflects better responses. 
\paragraph{\large Our Contribution:} In response to the aforementioned limitations, in this study, we aim to develop an improved counterspeech generation pipeline, one that produces responses that are both aligned to the desired intent while also attentive to the short and implied nature of the hate speech.  In total, we consider four counterspeech intents --\text{positive}, \text{informative}, \text{question}, and \text{denouncing}. We curate \texttt{IntentCONANv2}, the largest \text{intent-specific counterspeech generation dataset} consisting of $13,952$ counterspeeches for $3,488$ hate speech instances. Further, we propose \texttt{CoARL}, a {novel three-phased counterspeech generation framework}. In the first stage, \texttt{CoARL} learns to generate explanations along different pragmatic and implied dimensions of hate speech. In the second stage, \texttt{CoARL} is trained to generate intent-specific counterspeeches by learning task-specific adapters. Finally, we fine-tune a policy using reinforcement learning by designing a composite reward function to optimize the model's output towards being \textit{effective} and \textit{non-toxic}. An extensive comparison using automated and human evaluation suggests that our proposed method consistently beats the current counterspeech generation benchmarks across multiple metrics and shows comparative performance with LLMs like ChatGPT. 

%% file: sections/related_work.tex
\section{Related Work}
\paragraph{\large Automatic Counterspeech Generation:}
\citet{qian2019benchmark} made an initial attempt to generate counterspeeches with the seq2seq model. \citet{zhu-bhat-2021-generate} developed a three-task pipeline that includes an encoder, a grammar check, and counterspeech retrieval based on hate speech to produce diverse counterspeeches. While existing studies indicate the effectiveness of conditioned counterspeech based on context \cite{mathew2019thou, hangartner2021empathy}, effective counterspeech generation is still at its nascent stage. \citet{saha2019complex} introduced CounterGEDI, a model designed to control attributes like politeness, detoxification, and emotions in generated counterspeeches using class-conditioned language models. Recently, \citet{gupta-etal-2023-counterspeeches} proposed a two-phased pipeline to generate \textit{intent-specific } counterspeeches. 

\paragraph{\large Instruction Tuning and RLAIF:} 
Instruction tuning (IT) enhances the functionality and controllability of large language models (LLMs), yielding more predictable behaviors compared to standard LLMs \cite{wang2022supernaturalinstructions, mishra2022crosstask, zhang2023instruction}. Studies such as \cite{wei2022finetuned} demonstrate that multitask fine-tuning with instructions on moderate-size LMs facilitates zero-shot task generalization. This is achieved by scaling the number of training tasks, prompts per task, and LM size. IT has also proven effective in controlled text generation, often outperforming other methods in constraint satisfaction. \citet{zhou2023-InstructCTG} highlighted that incorporating conditions into instructions enhances controlled text generation. This approach allows models to dynamically adapt to constraints, improving task-specific generation and zero-shot constraint generalization. By verbalizing constraints in natural language, the prompt-based generation capabilities of pre-trained models are optimally utilized, enabling them to address new, unseen constraints during training by simply describing them in natural language. We use a similar strategy in our method where we verbalize the \textit{intent} as part of the instruction itself (c.f Table \ref{tab:instructions_table}).

Recently, IT combined with Reinforcement learning from human feedback has been efficacious in aligning LLMs with human preferences in various applications, including summarisation, dialogue, and question answering \cite{ouyang2022-instuctGPT, glaese2022improving}, with recent work introducing Reinforcement Learning from AI Feedback (RLAIF) \cite{lee2023rlaif} for optimizing helpfulness and harmlessness, and demonstrating that RLAIF can achieve improvements comparable to traditional methods without relying on human annotators \cite{ziegler2020finetuning, christiano2023deep, nakano2022webgpt, bai2022constitutional}. Our approach aims to explore the use of pretrained classifiers to align an instruction-tuned LLM towards certain desired attributes of e\textit{ffectiveness} and \textit{non-toxicity}. 

%% file: tables/table1.tex
\begin{table}[t!]
\begin{center}
\resizebox{\columnwidth}{!}{%
\begin{tabular}{l|c|c|c|c|c}
\toprule
\multicolumn{1}{l|}{\textbf{Hate speech}} & \multicolumn{5}{c}{\textbf{Counterspeech Intent}} \\ 
\midrule
\textbf{Target group} & \textbf{INF} & \textbf{POS} & \textbf{QUE} & \textbf{DEN} & \textbf{Total} \\ 
\midrule
\midrule
\textbf{Muslims} & 914 & 914 & 914 & 914 & 3656 \\
\textbf{Women} & 508 & 508 & 508 & 508 & 2032 \\
\textbf{LGBTQ+} & 449 & 449 & 449 & 449 & 1796 \\
\textbf{Jews} & 392 & 392 & 392 & 392 & 1568 \\
\textbf{Refugees} & 70 & 70 & 70 & 70 & 280 \\
\textbf{Asian people} & 29 & 29 & 29 & 29 & 116 \\
\textbf{Immigrants} & 562 & 562 & 562 & 562 & 2248 \\
\textbf{Disabled} & 173 & 173 & 173 & 173 & 304 \\
\textbf{POC} & 306 & 306 & 306 & 306 & 80 \\
\textbf{Other} & 85 & 85 & 85 & 85 & 208 \\
\midrule
\textbf{Total} & 3488 & 3488 & 3488 & 3488 & 13952 \\
\midrule
\textbf{Train} & 2383 & 2383 & 2383 & 2383 & 9532 \\
\textbf{Dev} & 365 & 365 & 365 & 365 & 1460 \\
\textbf{Test} & 740 & 740 & 740 & 740 & 2960 \\
\bottomrule
\end{tabular}%
}
\end{center}
\caption{Statistics of the \texttt{IntentCONANv2} dataset.} 
\label{tab:table1_stats}
\vspace{-5mm}
\end{table}

%% file: sections/dataset.tex
\section{Dataset}
We introduce \texttt{IntentCONANv2}, an expanded and refined version of the publicly-available \texttt{IntentCONAN} dataset \cite{gupta-etal-2023-counterspeeches}. \texttt{IntentCONANv2} is a large-scale dataset comprising $13,952$ \texttt{CS} instances across four distinct intents: positive (POS), informative (INF), questioning (QUE), and denouncing (DEN). The development of \texttt{IntentCONANv2} involved addressing key limitations of the original dataset, leading to significant improvements in both content and structure.
Firstly, we eliminated the \texttt{humor} intent, acknowledging its subjective nature and tendency to produce vague or offensive content \cite{chung-et-al-2023-understanding-counterspeech, gupta-etal-2023-counterspeeches}. Secondly, we addressed the non-uniform distribution of counterspeeches in the original dataset (See Table \ref{tab:table1_stats}). While \texttt{IntentCONAN} had an inconsistent representation of counterspeeches across different hate speech instances, \texttt{IntentCONANv2} ensures an average of four counterspeeches per hate speech, significantly improving upon the original average of two. Further improvements include a focus on the length and content quality of the counterspeeches. The effectiveness of a robust counterspeech is reflected in its length, with a greater token count indicating a more comprehensive response. A higher token count suggests that the counterspeech encompasses a broader range of information to effectively counteract hate speech. Our analyses indicate a notable presence of overly brief responses in the original dataset, particularly in the denouncing and questioning intents. To address this, \texttt{IntentCONANv2} emphasizes generating counterspeeches with substantial content, increasing the average token length from $26.48$ to $40.61$.
Appendix \ref{sec:appendix-dataset} shows more details about the dataset and annotation process.

%% file: figures/figure2.tex
\begin{figure*}[t]
\includegraphics[width=\textwidth]{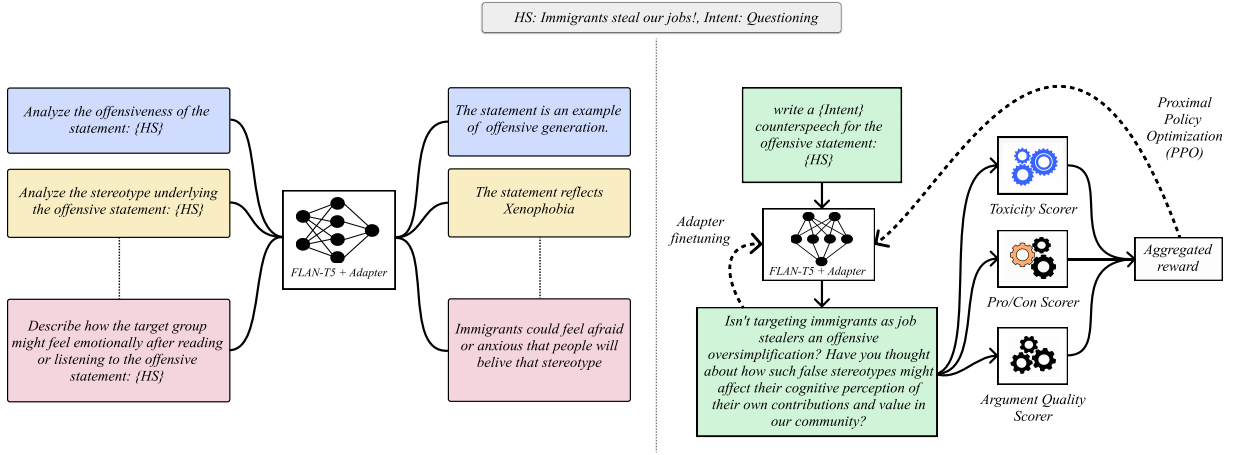}
\caption{Overview of the three-phased architecture of \texttt{CoARL}. In the first phase (left), \texttt{CoARL} is trained on an auxiliary task of hate speech (\texttt{HS}) explanation generation using a multi-task IT setup. Subsequently, in the second phase (right), task-specific LoRA weights are trained by freezing the model parameters from the previous phase, thus, enabling forward knowledge transfer without catastrophic forgetting. In the final phase (right), the model output is optimized via RL using feedback from a composite reward model consisting of three pre-trained classifiers.}
\label{fig:figure2}
\end{figure*}

%% file: sections/proposed_methodology.tex
\section{Proposed Methodology}
\label{sec:proposed_methodology}
In this section, we describe \texttt{\texttt{CoARL}}, a novel framework for automated counterspeech generation designed to address two main challenges: (i) generating intent-specific counterspeech that is both topically and semantically \textit{relevant} to the hate speech, and (ii) aligning the counterspeech with the desired values of  \textit{effectiveness} and \textit{non-toxicity} (Fig. \ref{fig:figure2}). 

\subsection*{Task Formulation}
Let us denote the \texttt{Intentconan2} dataset as  \( \mathcal{D} = \{ (x_1, c_1, y_1), \ldots, (x_n, c_n, y_n) \} \), where \( x_i \in \mathcal{X} \) is the \( i \)-th hate speech statement, \( y_i \in \mathcal{Y} \) is the counterspeech corresponding to \( x_i \), and \( c_i \in \mathcal{C} \) is the category/intent of \( y_i \). We aim to learn a stochastic counterspeech generation function \( \chi : \mathcal{X} \times \mathcal{C} \rightarrow \mathcal{Y} \), such that \( y_i \sim \chi(\cdot|x_i, c_i) \).

We approach this problem by decomposing the counterspeech generation task into three phases. In the first phase, we train a base language model on seven hate speech explanation tasks, each covering a unique pragmatic facet of hate speech. In the second phase, we freeze the model parameters learnt during the previous phase, and fine-tune a task-specific  Low-Rank Adapter (LoRA) for counterspeech generation. Finally, we train a policy using reinforcement learning to optimize the model's output to be both effective and non-toxic simultaneously. We use FLAN-T5 \cite{chung2022scaling} as the base model for all our experiments. Note that while there is a range of valid model choices when it comes to seq2seq modelling, we choose FLAN-T5 based on its strong reasoning abilities \cite{chung2022scaling}.

\subsection*{Phase 1: Auxiliary Explanation Generation}
\label{sec:phase1} 
Following recent work in hate speech explanation generation, we make use of \texttt{COBRACORPUS} \cite{zhou-etal-2023-cobra}, a dataset of offensive statements paired with free-text explanations along seven pragmatic frames of hate speech $-$ intent, target group, power-dynamics, impact, emotional reaction, cognitive reaction, and offensiveness. Contrary to \citet{zhou-etal-2023-cobra} who generate explanations in a linearized format, we adopt a multi-task instruction tuning setup (c.f. Fig \ref{fig:figure2}). 

Let $\pi^{PRE}$ be a vanilla FLAN-T5 model parameterized by $\Theta$. We begin by converting \texttt{COBRACORPUS} into a set of independent instruction-tuning tasks along each of the seven explanation dimensions $\{I_1, I_2,\ldots,I_7\}$  (See Table \ref{tab:instructions_table}). Following this, we fine-tune $\pi^{PRE}$ simultaneously for each of the seven explanation generation tasks, with the aim of learning a shared representation that will enable the model to generalize better on each task. The common multi-task training objective for all tasks $(N = 7)$ can be written as follows: 
\begin{equation}\small
\begin{aligned}
\Theta_{m} = \underset{\Theta}{\mathrm{argmin}} \sum_{n=1}^{N} L_n(I_n; \Theta)
\end{aligned}
\end{equation}
where $I_n$ corresponds to the instruction set for task $t \in \{1,...,N\}$, $L_0$ denotes standard cross-entropy loss,  $\Theta_0 \rightarrow \{1,...,N\}$ indicates that we start with $\Theta_0$ and fine-tune on a set of tasks $\{1, ..., N\}$, and $\Theta_m$ are the parameters of the fine-tuned model.

\subsection*{Phase 2: Task-Specific Adapter Learning} 
\label{sec:phase2} 
We formulate the task of intent-conditioned counterspeech generation, in which we verbalize the intent as part of the instruction itself (See ${I_8}$, Table \ref{tab:instructions_table}). Then, instead of fine-tuning a new model from scratch, we train task-specific LoRA adapter weights \cite{hu2021lora} on top of the model parameters learned during the previous phase. This setup provides two benefits. First, it allows for forward knowledge transfer where the model is able to leverage the knowledge learnt during explanation generation from phase 1. Second, learning task-specific LoRA parameters independently helps avoid catastrophic forgetting. 

We initialize a FLAN-T5 model with \textbf{$\Theta_m$}, i.e., the parameters learned from the previous phase. We then freeze the model's parameters and fine-tune task-specific LoRA parameters for intent-conditioned counterspeech generation. Specifically, we apply LoRA to the query and value projection matrices in the self-attention module of FLAN-T5, following \citet{hu2021lora}. Let $W_m$ denote the weight matrix corresponding to the model initialized with $\Theta_m$. For each pre-trained weight matrix $W_m \in \mathbb{R}^{d \times d}$, we introduce two trainable matrices $A \in \mathbb{R}^{r \times d}$ and $B \in \mathbb{R}^{d \times r}$, where $r \ll d$ is the rank of LoRA. We constrain the update to the weight matrix by $W_m + \Delta W = W_m + BA$, where $\Delta W$ is the low-rank adaptation. We initialize $A$ randomly and $B$ to zero, and scale $\Delta W x$ by $\alpha / r$, where $\alpha$ is a constant and $x$ is the input vector. We freeze $W_m$ and only optimize $A$ and $B$ using the Adam optimizer. Finally, we fine-tune the LoRA parameters on $\mathcal{D}$ by maximizing the log-likelihood of the target given the source instruction:
\begin{equation}\small
\begin{aligned}
\max_{A, B} \sum_{(x, y) \in \mathcal{D}} \sum_{t=1}^{|y|} \log (\pi^{PRE}(y_t | x, y_{<t}; A, B))
\end{aligned}
\end{equation}
where $x$ is the hate speech instance, $y$ is the target counterspeech, and $y_t$ is the $t$-th token in $y$. We use the cross-entropy loss as the objective function. 

\subsection*{Phase 3: Optimization via Reinforcement Learning}
\label{sec:phase3} 
In this phase, we optimize our model to generate counterspeeches that are both effective and non-toxic, using a composite reward function that combines three types of rewards: stance (pro-con), argument quality, and toxicity. Inspired by \citet{bai2022constitutional}, our proposed RLAIF pipeline consists of three phases: supervised fine-tuning, reward modeling, and reinforcement learning-based fine-tuning.

\paragraph{Supervised Fine-Tuning:} Let $\pi^{SFT}$ parameterized by $\theta_{SFT}$
denote the supervised fine-tuned model learned during phase 2. Specifically, $\pi^{SFT}$ can be represented as
$\theta_{SFT} = W_m + BA$,
where $W_m \in \mathbb{R}^{d \times d}$ is the weight matrix learned in phase 1 (for the task of hate speech explanations generation), and $A \in \mathbb{R}^{r \times d}$ and $B \in \mathbb{R}^{d \times r}$, are the two low-rank LoRA matrices learned during phase 2 (for the task of intent-specific counterspeech generation).

\paragraph{Reward Model (RM):}
We use three transformer-based models, each trained on the tasks of \textbf{stance (pro-con / PC)} classification, \textbf{argument quality (AQ)} prediction, and \textbf{toxicity (T)} prediction to provide the reward signals for each generated counterspeech (see Appendix \ref{sec:appendix-reward-model} for details). The reward function aims to encourage the model to generate counterspeeches that contradict hate speech, form logical and persuasive arguments, and avoid harmful or offensive language. The overall reward is computed by taking the mean of these components after normalizing them to a unified scale of 0 to 1, where 1 represents the ideal reward. The reward function is formulated as:
\begin{equation}\small
\begin{aligned}
r(x', y') = \frac {1}{3} \left(\frac{1 - PC(x', y')}{2}\right) + AQ(y') + \\ \left(1 - T(y')\right)
\end{aligned}
\end{equation}

In this context, \(x'\) denotes instruction-formatted hate speech, while \(y'\) denotes the counterspeech generated by  \(\pi^{SFT}\). 
The Pro-Con score \(PC(x',y')\) is normalized to \(\frac{1 - PC(x', y')}{2}\) to convert the original scale of [-1, 1] to [0, 1], thereby aligning lower scores with higher rewards. Additionally, \(AQ(y')\) represents the argument quality score, which is already within the [0, 1] range. Finally, \(T(y')\) is inverted to \(1 - T(y')\) to transform the original scale of [0, 1] (where lower is better) to a scale where higher scores indicate lower toxicity. By averaging these normalized components, our RM incentivizes the generation of counterspeech that is not only relevant and persuasive but also minimizes toxicity.

\paragraph{Reinforcement Learning:}
We use the Proximal Policy Optimization (PPO) algorithm \cite{schulman2017proximal}. We initialize a policy model $\pi^{RL}$ from the SFT model $\pi^{SFT}$ and fine-tune it using the reward signals given by the reward model RM. We also add a regularization term in the form of KL divergence to the final objective function, to ensure a smooth and natural gradient update and prevent the policy from deviating too much from the original model.
\begin{equation}\small
\begin{aligned}
\mathcal{L}_{PPO}(\pi^{RL}) = & \mathbb{E}_{x \sim D, y \sim \pi^{RL}(x)}\bigg[ \frac{\pi^{RL}(y|x)}{\pi^{SFT}(y|x)}r(y) \\
& - \epsilon \cdot KL(\pi^{SFT}(x) || \pi^{RL}(x)) \bigg] \\
\end{aligned}
\end{equation}
where $\pi^{RL}(y|x)$ and $\pi^{SFT}(y|x)$ are the probabilities of generating counterspeech $y$ given hate speech $x$ by the policy model $\pi^{RL}$ and the SFT model $\pi^{SFT}$, respectively, $r(y)$ is the reward given by the reward model for counterspeech $y$, and $KL(\pi^{SFT}(x) || \pi^{RL}(x))$ is the Kullback-Leibler Divergence (KL) between the probability distributions of counterspeeches generated by $\pi^{SFT}$ and $\pi^{RL}$ for a given hate speech $x$.
\begin{equation}\small
\begin{aligned}
\scalebox{0.95}{%
$KL(\pi^{SFT}(x) || \pi^{RL}(x)) = \sum_{x} P(x) \log \frac{\pi^{SFT}(x)}{\pi^{RL}(x)}$
}
\end{aligned}
\end{equation}
The KL term is used to penalize large changes in the policy and ensure a smooth update. The hyperparameter $\epsilon$ controls the trade-off between exploration and exploitation. Thus, the final objective function aims to maximize the expected reward while keeping the policy model $\pi^{RL}$ close to the SFT model $\pi^{SFT}$.

%% file: tables/results_table.tex
\begin{table*}[!ht]
\begin{center}
\resizebox{\textwidth}{!}{%
\begin{tabular}{lccccccccccc} 
\toprule
\textbf{Method} & \textbf{Prompt/Adapter} & \multicolumn{3}{c}{\textbf{ROUGE} (↑)} & \textbf{M} (↑) & \textbf{BS} (↑) & \textbf{CS} (↑) & \textbf{CA} (↑) & \textbf{PC} (↓) & \textbf{AQ} (↑) & \textbf{T} (↓) \\ 
\cmidrule(lr){3-5}
~ & ~ & \textbf{R1} & \textbf{R2} & \textbf{RL} & ~ & ~ & ~ & ~ & ~ & ~ & ~ \\ 
\midrule
GPS & $-$ & $0.175$ & $0.026$ & $0.151$ & $0.128$ & $0.856$ & $0.150$ & $0.295$ & $-0.013$ & $0.679$ & $0.126$ \\ 
DialoGPT & $-$ & $0.236$ & $0.061$ & $0.208$ & $0.230$ & $0.875$ & $0.202$ & $0.921$ & $-0.118$ & $0.811$ & $0.106$ \\ 
QUARK & $-$ & $0.219$ & $0.062$ & $0.191$ & $0.174$ & $0.874$ & $0.182$ & $0.745$ & $-0.030$ & $0.790$ & $0.108$ \\ 
\midrule
Vanilla FLAN-T5$_{\mathrm{XXL}}$ & ZS & $0.175$ & $0.042$ & $0.157$ & $0.123$ & $0.859$ & $0.148$ & $0.528$ & $-0.113$ & $0.710$ & $0.321$ \\ 
Vanilla FLAN-T5$_{\mathrm{XXL}}$ & FS & $0.177$ & $0.043$ & $0.158$ & $0.125$ & $0.869$ & $0.148$ & $0.509$ & $-0.120$ & $0.705$ & $0.299$ \\ 
AEG FLAN-T5$_{\mathrm{XXL}}$ & ZS & $0.185$ & $0.041$ & $0.169$ & $0.126$ & $0.873$ & $0.161$ & $0.518$ & $-0.082$ & $0.730$ & $0.263$ \\ 
AEG FLAN-T5$_{\mathrm{XXL}}$ & FS & $0.184$ & $0.043$ & $0.165$ & $0.126$ & $0.870$ & $0.167$ & $0.517$ & $-0.125$ & $0.728$ & $0.268$ \\ 
\midrule
GPT-3.5-Turbo & ZS & $0.204$ & $0.058$ & $0.181$ & $0.274$ & $0.856$ & $0.323$ & $0.828$ & $0.118$ & $0.898$ & \underline{$\boldsymbol{0.038}$} \\ 
GPT-3.5-Turbo & FS & $0.230$ & $0.067$ & $0.199$ & \underline{$\boldsymbol{0.293}$} & \underline{$\boldsymbol{0.885}$} & \underline{$\boldsymbol{0.310}$} & $0.891$ & $-0.045$ & \underline{$\boldsymbol{0.914}$} & $0.043$ \\ 
\midrule
GPT-4-Turbo & ZS & $0.204$ & $0.058$ & $0.181$ & $0.274$ & $0.856$ & $0.323$ & $0.828$ & $0.118$ & $0.898$ & \underline{$\boldsymbol{0.038}$} \\ 
GPT-4-Turbo & FS & $0.230$ & $0.067$ & $0.199$ & \underline{$\boldsymbol{0.293}$} & \underline{$\boldsymbol{0.885}$} & \underline{$\boldsymbol{0.310}$} & $0.891$ & $-0.045$ & \underline{$\boldsymbol{0.914}$} & $0.043$ \\ 
\midrule

\textbf{CoARL (Ours)} & LoRA16 & \underline{$\boldsymbol{0.251}$} & \underline{$\boldsymbol{0.078}$} & \underline{$\boldsymbol{0.221}$} & $0.244$ & $0.876$ & $0.226$ & \underline{$\boldsymbol{0.944}$} & \underline{$\boldsymbol{-0.130}$} & $0.824$ & $0.067$ \\ 
\midrule
- RL & LoRA16 & $0.251$ & $0.071$ & $0.220$ & $0.249$ & $0.868$ & $0.231$ & $0.946$ & $-0.112$ & $0.815$ & $0.101$ \\ 
- reward (Toxicity) & LoRA16 & $0.251$ & $0.078$ & $0.220$ & $0.244$ & $0.874$ & $0.226$ & $0.943$ & $-0.130$ & $0.823$ & $0.107$ \\ 
- reward (AQ) & LoRA16 & $0.248$ & $0.076$ & $0.217$ & $0.232$ & $0.868$ & $0.223$ & $0.938$ & $-0.129$ & $0.804$ & $0.076$ \\ 
- reward (PC) & LoRA16 & $0.245$ & $0.076$ & $0.215$ & $0.239$ & $0.865$ & $0.221$ & $0.937$ & $-0.107$ & $0.815$ & $0.071$ \\ 
- AEG & $-$ & $0.247$ & $0.069$ & $0.216$ & $0.245$ & $0.862$ & $0.222$ & $0.930$ & $-0.113$ & $0.816$ & $0.124$ \\ 
- LoRA (SFT) & $-$ & $0.234$ & $0.067$ & $0.210$ & $0.233$ & $0.809$ & $0.215$ & $0.939$ & $-0.111$ & $0.801$ & $0.106$ \\ 
\midrule
\multicolumn{2}{c}{$\Delta_{\mathrm{CoARL (Ours)} - \mathrm{Best Method}}$} & \textcolor{ForestGreen}{$\uparrow0.015$} & \textcolor{ForestGreen}{$\uparrow0.011$} & \textcolor{ForestGreen}{$\uparrow0.012$} & \textcolor{red}{$\downarrow0.049$} & \textcolor{red}{$\downarrow0.009$} & \textcolor{red}{$\downarrow0.097$} & \textcolor{ForestGreen}{$\uparrow0.024$} & \textcolor{ForestGreen}{$\uparrow0.005$} & \textcolor{red}{$\downarrow0.090$} & \textcolor{red}{$\downarrow0.090$} \\
\bottomrule
\end{tabular}%
}

\end{center}
\caption{Comparative evaluation of \texttt{CoARL} against state-of-the-art models across multiple evaluation metrics. The symbol ↑ (↓) indicates the higher (lower) value is better. }
\label{tab:results_table}
\end{table*}

%% file: sections/experimental_setup.tex
\section{Experimental Setup}
\subsection{Baselines}
\label{sec:baselines}
We report \textbf{Generate Prune Select (GPS)} by \cite{zhu-bhat-2021-generate}, which employs a three-stage pipeline including an autoencoder for initial counterspeech generation, a grammatical pruning model, and a vector-based response selection model. We also fine-tune \textbf{DialoGPT} \cite{zhang2020dialogpt} for its ability to generate contextually relevant responses, surpassing similar models like GPT-2. Additionally, we include \textbf{QUARC} \cite{gupta-etal-2023-counterspeeches}, recognized as the current state-of-the-art in intent-conditioned counterspeech generation. 
To further broaden our comparison scope, we also consider prompting baselines leveraging the capabilities of LLMs. Recent developments in in-context learning have revealed that these models can achieve performances comparable to, or even surpass, traditional supervised fine-tuning across various NLP tasks. Therefore, for a comprehensive evaluation, we report both zero-shot and few-shot performances on three LLMs -- \textbf{Vanilla FLAN-T5$_{\mathrm{XXL}}$} \cite{chung2022scaling},  \textbf{AEG FLAN-T5$_{\mathrm{XXL}}$}, i.e., FLAN-T5$_{\mathrm{XXL}}$ trained on auxiliary explanation generation, and ChatGPT (\textbf{GPT-3.5-Turbo}) \cite{ouyang2022-instuctGPT}. Detailed methodologies and results of these prompting experiments are presented in Appendix \ref{sec:appendix-prompting}.

\subsection{Evaluation Metrics}
Evaluating counterspeech generation presents unique challenges due to the task's inherent open-ended nature, varied correct responses, and the absence of standard evaluation criteria \cite{chung-et-al-2023-understanding-counterspeech}. To address these challenges, our evaluation adopts a multidimensional approach, each focusing on a specific aspect of counterspeech quality. These dimensions include lexical similarity, relevance, effectiveness, intent conformity, and toxicity.
\textit{Lexical similarity}, measuring the linguistic alignment between generated and reference counterspeech, is evaluated using \textbf{Rouge} \cite{rouge-score-lin-2004} and \textbf{Meteor} \cite{meteor-score-banerjee-lavie-2005}. \textit{Relevance}, focusing on the counterspeech's direct engagement with the primary topic of the hate speech, is assessed through cosine similarity \cite{sentence-transformers} and \textbf{BERTScore} \cite{bert-score-zhang2020}, ensuring topical and semantic coherence. A low relevance score suggests a topic mismatch, where the counterspeech diverges from the primary subject of the hate speech. \textit{Effectiveness} is assessed using Project Debater’s API services: \textbf{Pro/Con (PC)} and \textbf{Argument Quality (AQ)} \cite{bar-haim-etal-2021-project-debator}. The \textbf{PC} metric evaluates whether an argument supports or opposes a given topic, while \textbf{AQ} assigns a quality score to the argument. We treat hate speech as the topic and the generated counterspeech as the argument, calculating these metrics for each model output. \textbf{Category Accuracy (CA)}, following \cite{gupta-etal-2023-counterspeeches}, measures how effectively each model incorporates the intended intent into the generated counterspeech. Finally, we evaluate \textbf{\textit{Toxicity}}\footnote{\href{3}{https://www.perspectiveapi.com/}} using the Detoxify library, an unbiased toxicity classification model \cite{Detoxify}, to ensure that the generated counterspeech does not perpetuate harmful language.

%% file: sections/results_and_discussion.tex
\section{Experimental Results}
In this section, we delve into a comprehensive analysis of the experimental results, comparing the performance of our proposed method, {\texttt{CoARL} (Ours)} against state-of-the-art baselines.

\subsection{Quantitative Results}
Table \ref{tab:results_table} shows quantitative results across a spectrum of automated metrics, illustrating the superior performance of \texttt{CoARL} in various evaluative criteria. In ROUGE-based scores, namely R1, R2, and RL, \texttt{CoARL} achieves outstanding scores of $0.251$, $0.078$, and $0.221$, respectively, demonstrating its effectiveness in generating lexically-aligned counterspeech, a key factor for relevance and appropriateness in responding to hate speech. Furthermore, \texttt{CoARL} attains an impressive Category Accuracy (CA) score of $0.944$, underscoring its precision in incorporating the intended intent within counterspeech. When compared to baseline models such as GPS, DialoGPT, and QUARK, \texttt{CoARL} not only excels in lexical similarity but also in relevance and argument quality. This is evidenced by its higher BERTScore ($0.876$) and Argument Quality ($0.824$) while maintaining a significantly lower toxicity score ($0.067$). 

\texttt{CoARL}'s outputs also exhibit higher semantic and topical relevance, indicated by its Meteor, BERTScore, and Cosine Similarity scores, only surpassed by GPT-3.5, which benefits from its larger model architecture and RLAIF fine-tuning. This comparison highlights the importance of the model size and fine-tuning approaches in counterspeech generation. The inclusion of Auxiliary Explanation Generation (AEG) in our methodology has proven beneficial, as FLAN-T5 models trained with AEG outperform their vanilla counterparts in lexical and semantic similarity metrics, in both zero- and few-shot settings, further validating the hypothesis that hate speech explanation generation enhances counterspeech quality. Notably, \texttt{CoARL} outperforms GPT-3.5 in CA and PC metrics, emphasizing its ability to generate counterspeech that effectively counters hate speech while maintaining comparable performance in toxicity scores, thus generating impactful yet non-toxic counterspeech.

\subsection{Ablation Study}
This section details ablation experiments conducted to evaluate the significance of various components within the \texttt{CoARL} framework. We assess CoARL's performance with and without Reinforcement Learning (RL), Auxiliary Explanation Generation (AEG), the LoRA adapter, and using different reward functions, as presented in Table \ref{tab:results_table}. The results demonstrate a performance decline upon removing any of these components, underscoring their collective importance.
$\bullet$ RL optimizes counterspeech for effectiveness and non-toxicity using a composite reward function (stance, argument quality, and toxicity). Without RL, counterspeech effectiveness diminishes, indicated by lower Pro/Con and Argument Quality scores, and toxicity increases. Each reward component uniquely influences counterspeech quality. Excluding the stance reward slightly decreases Pro/Con scores but increases Argument Quality, implying some counterspeeches might be persuasive without directly opposing hate speech. Removing argument quality rewards significantly lowers Argument Quality scores but slightly decreases Toxicity, suggesting possible harmful language in high-quality arguments. Eliminating the toxicity reward notably raises Toxicity scores but slightly improves Pro/Con and Argument Quality, indicating some effective counterspeeches might use offensive language.
{\Large•} AEG enhances the model's understanding of hate speech context and nuances, leading to more relevant and coherent counterspeeches. Without AEG, ROUGE, Meteor, BERTScore, Cosine Similarity, Argument Quality, and Category Accuracy scores drop.
{\Large•} The LoRA adapter allows for learning task-specific parameters in counterspeech generation without erasing prior knowledge. In its absence, the model uses sequential fine-tuning of the base FLAN-T5 model, resulting in reduced ROUGE, BERTScore, Argument Quality, and Category Accuracy, along with an increase in Toxicity.

\subsection{Human Evaluation}
Given the open-ended nature of the problem, we undertook an extensive human evaluation. We analyze a randomly selected subset of counterspeeches generated by three best-performing methods from our quantitative evaluation (See Table \ref{tab:results_table}): \texttt{COARL}, the SFT model (\texttt{COARL} - RLAIF), and few-shot ChatGPT (GPT-3.5 Turbo). The subset was uniformly distributed across all four intents. For a given hatespeech, we ask our evaluators\footnote{The evaluation panel consisted of 35 experts from the fields of NLP and social science, aged between 20-30 years, with a gender distribution of 45\% male and 55\% female.} to rank responses from the each of the aforementioned models across the following metrics:
{\Large•} \textbf{Independent Counterspeech (IC)} evaluates the ability of the generated counterspeech to function independently, without reliance on additional context.
{\Large•} \textbf{Adequacy (A)} is used to assess the grammatical correctness, coherence, and fluency of the counterspeech, examining how effectively it adheres to grammatical norms and syntactical clarity.
{\Large•} \textbf{Contextual Relevance (CR)} denotes the counterspeech’s capacity to address key elements of hate speech, including subject matter, false claims, targeted group, projected stereotypes, or biases.
{\Large•} \textbf{Argumentative Effectiveness (AE)} measures the counterspeech's success in presenting cogent and convincing arguments in response to hate speech. High AE is indicative of a logically structured and impactful counterargument.
{\Large•} \textbf{Category Accuracy (CA):} This evaluates the extent to which the counterspeech aligns with its intended objective, based on the categorization of its intent by evaluators.
For each of these metrics, we report the \textit{Win Rate} of \texttt{CoARL} against the SFT model and ChatGPT in \ref{tab:human_eval}. \textit{Win Rate} evaluates the end-to-end quality of two models, measuring how often one model is preferred by humans over the other. The percentage of instances where model \textsc{A} is preferred over model \textsc{B} is referred to as \textit{"Win Rate of A vs B"}.

%% file: tables/human_eval.tex
\begin{table}[t!]
\begin{center}
\resizebox{\columnwidth}{!}{%
\begin{tabular}{l|c|c|c|c|c}
\toprule
\multicolumn{1}{l|}{\textbf{Model}} & \multicolumn{5}{c}{\textbf{Human Evaluation Metric}} \\
\midrule
 & \textbf{IC $\uparrow$} & \textbf{A $\uparrow$} & \textbf{CR $\uparrow$} & \textbf{AE $\uparrow$} & \textbf{CA $\uparrow$} \\
\midrule
\texttt{CoARL} vs SFT & $\textbf{74}$ & $\textbf{60}$ & $\textbf{72}$ & $\textbf{81}$ & $\textbf{68}$ \\
\texttt{CoARL} vs ChatGPT & $\textbf{58}$ & ${56}$ & $\textbf{59}$ & ${62}$ & $\textbf{74}$ \\
\bottomrule
\end{tabular}%
}
\end{center}
\caption{Results for different human evaluation metrics. We report\textbf{ Win Rate \%} for responses generated from \texttt{CoARL} against those generated by a) SFT model (\texttt{CoaRL} - RL) and ChatGPT (GPT-3.5 Turbo).}
\label{tab:human_eval}
\vspace{-5mm}
\end{table}

%% file: sections/conclusion.tex
\section{Conclusion}
To address online hate speech with effective and diverse responses, we introduced the task of automated counterspeech generation, integrating pragmatic reasoning. We created \texttt{CoARL}, a three-stage framework that leverages instruction tuning and reinforcement learning to generate high-quality counterspeeches that are aligned to the pragmatic facets and intents of hate speech. We also presented {\tt IntentCONANv2},  consisting of $13,952$ intent-specific counterspeeches. We performed a comprehensive evaluation across both quantitative and qualitative metrics to demonstrate the superiority of \texttt{CoARL} over existing methods and ablations.

\section*{Acknowledgement}
The work was financially supported by Logically.

%% file: sections/limitation.tex
\section*{Limitation}
Our study has limitations that should be acknowledged. First, our dataset of hate speech and counterspeech is not exhaustive, and may not cover all possible types and targets of online hate. Second, our framework relies on pre-trained models for reward modeling and reinforcement learning, which may introduce biases or errors from the source models. Third, our evaluation metrics are not fully aligned with the human perception of counterspeech quality, and may not capture the nuances and subtleties of natural language. Fourth, our framework does not account for the potential feedback loop or escalation that may occur after generating counterspeech, which may affect the long-term effectiveness and impact of our approach. Future work could address these limitations by expanding and diversifying the dataset, improving the reward function and evaluation criteria, and incorporating human feedback and dialogue modeling into the framework.

%% file: sections/ethics_statement.tex
\section*{Ethics Statement}
We acknowledge the sensitivity involved in addressing online hate speech and understand that conducting research in this area may lead to ethical and moral dilemmas. This project marks an initial step towards developing a comprehensive and varied collection of counterspeech responses for each instance of hate speech. We recognize that automated counterspeech models might generate inaccurate statements, highlighting the need for a more effective integration of real-world knowledge within these models. Despite the potential proficiency of generative models, the urgent requirement for an extensive, varied counterspeech dataset to achieve overall positive results remains. Moreover, while fully operational counterspeech models are not yet a reality, organizations like United Against Hate are significantly contributing to reducing hate speech online.

%% file: sections/appendix.tex
\section{Dataset Analysis}
\label{sec:appendix-dataset}


\paragraph{Annotation Process and Guidelines:}
Before initiating the annotation process, comprehensive familiarization with the field manual\footnote{\url{https://onlineharassmentfieldmanual.pen.org/}} on ``responding to online abuse'' was ensured for all annotators. In the preparatory phase of our study, we engaged in multiple discussions with the annotators to enhance their understanding of counterspeech. The annotators were directed to focus on several pivotal goals for each type of counterspeech:

\begin{enumerate}
    \item \textbf{Defining the Purpose:} Every counterspeech variant should embody a unique core concept, speech style, and intended outcome.
    \item \textbf{Reducing Conflict:} The aim was to ensure that each counterspeech would mitigate the situation without causing further escalation or eliciting additional hate speech.
    \item \textbf{Eliminating Aggressive Language:} There was a strict prohibition against the use of aggressive language, including threats, name-calling, and profanity in the counterspeeches.
\end{enumerate}

Following these guidelines, the annotators proceeded to write intent-specific counterspeech responses for a total of $3,488$ unique instances of hate speech. For a detailed breakdown of the data, refer to Table \ref{tab:table1_stats} under the IntentCONAN statistics section.

\paragraph{Statistical Analysis:}   In Figure \ref{4_all}, we present a detailed statistical analysis of \texttt{IntentCONANv2}. Our focus is on fine-graining the target groups into 10 distinct categories, a classification that significantly enhances the model's capacity to comprehend hate speech context in conjunction with intent. Figure \ref{4_1} specifically features a donut chart illustrating the distribution of hate speech target groups. To delve into specifics, we meticulously extracted refugees from the broader category "Immigrants," and this subgroup constitutes a noteworthy $2\%$ of the entire dataset. Simultaneously, we honed in on the "Asian people (AP)" category within the expansive classification of "Others," contributing a modest yet discernible $0.8\%$ to the overall dataset. Moving on to Figure \ref{4_2}, it provides the distribution of counter speech across the intent dimension. This multi-faceted examination contributes to a comprehensive evaluation of both hate speech and counterspeech dynamics within the dataset.

To delve into details regarding our expanded dataset, Figure \ref{4_3} provides insights into the uniform distribution of data points across training, validation, and testing sets. Examining Figure \ref{4_4} along the target group dimension reveals a consistent dispersion of counter speeches across different intents. Figures \ref{4_5} and \ref{4_6} elucidate the mean token distribution of counter speeches, both across intents and within various target groups. Notably, the counter speeches with the highest mean token length tend to lean towards positive intents, as evidenced by the findings. Simultaneously, a pattern of uniformity emerges in the mean token length distribution across the diverse target groups, contributing to a comprehensive understanding of the dataset characteristics.

\input{figures/dataset_analysis_appendix}
\input{tables/dataset_examples}
\section{Hate Speech Explanations}
\label{sec:appendix-hs-exp}
In this study, we employ the \texttt{COBRACORPUS} dataset \cite{zhou-etal-2023-cobra} to develop seven distinct subsets for instruction tuning. Each subset corresponds to specific explanation dimensions as outlined in Table \ref{tab:instructions_table}. We adhere to the original training, validation, and testing partitions as established in the COBRACORPUS dataset for each of these subsets. The explanation dimensions we incorporate are grounded in theoretical frameworks from pragmatics and implicature  \cite{grice1975logic,perez2021verbal} as well as the social psychology of bias and inequality \cite{nieto2006understanding,nadal2014emotional}. This approach allows for an extensive expansion of reasoning dimensions compared to previous studies, which primarily focused on identifying targeted groups and biased implications \cite{sap2019social,elsherief2021latent}. Below, we provide a comprehensive description of each explanation dimension.

\paragraph{Speaker Intent:} Speaker Intent captures the underlying communicative intent behind a hate speech statement (e.g., “to provoke” / "to make a joke" / "to insult"). Prior work has shown that intent can influence pragmatic implications related to biases and harms and aid in hate speech detection \cite{kasper1990linguistic,dynel2015landscape, holgate2018swear}.

\paragraph{Target Group:} Target Group describes the social or demographic group referenced or targeted by the hate speech statement (e.g., “those \textbf{jews}”, “hate \textbf{black} artists”), which could include the listener if they are targeted. This dimension has been the focus of several prior works as it is crucial towards understanding the offensiveness and harms of the statement \cite{zampieri2019predicting,sap2019social,vidgen2019how}.

\paragraph{Power Dynamics:} The Power Dynamics capture the socio-cultural power differential or axis of privilege-discrimination between the speaker and the target group or listener of a statement \cite{nieto2006understanding}. For example, a statement made by a white person to a black person may involve a racial power differential, where the white person has more privilege and social status than the black person. The power dimension helps to explain how the context affects the offensiveness and harm of a statement, as statements made by a more powerful speaker to a less powerful listener or target group may be more offensive or harmful than the other way around. The Power dimension is represented as a free-text description of the type of power dynamic involved in the statement, such as “gender differential”, “racial power differential”, or “class differential”.

\paragraph{Implication:} Implication or Implied Meaning explains the biased, prejudiced, or stereotypical meaning \textit{implied by the hate speech statement}, similar to \cite{sap2019social}. This implication is very closely related to the received meaning from the listener’s or targeted group’s perspective and may differ from the speaker’s intended meaning (e.g., for microaggressions). The Implication dimension is represented as a free-text description of the type of harm or effect involved in the statement, such as “emotional harm”, “stereotyping”, or “misinformation”. The Imp dimension is important for understanding the consequences of a statement and its potential to cause harm or offense. It is also useful for identifying the type of harm or effect that a statement can cause, which can help in designing effective interventions to mitigate the harms of toxic language.

\paragraph{Emotional Reaction:}
Emotional Reaction captures the possible negative effects and harms of the statement and its implied meaning on the targeted group or listener \cite{nadal2014emotional}. It is a free-text description of the short-term emotional effects or reactions to a statement (e.g., “anger and annoyance”, “worthlessness”). For example, in the statement "Wow, your English is really good!" uttered by a white person to a non-white person, the emotional reaction could be “listener and non-white people could feel angry, feel condescended to, or less confident about their own English skills”. This dimension helps to understand the psychological impact of offensive language and how it affects the well-being of the target group or listener.

\paragraph{Cognitive Reaction:}
Cognitive Reaction captures the possible cognitive effects and harms that the statement and its implied meaning could have on the targeted group or listener \cite{nadal2014emotional}. The cognitive reaction is a free-text description of the short-term cognitive effects or reactions to a statement (e.g., “confusion”, “doubt”, “disbelief”). For example, in the statement, “You are too young to understand this”, the cognitive reaction could be “the listener could feel confused, doubt their own understanding, or feel underestimated”. This dimension helps to understand the cognitive impact of offensive language and how it affects the well-being of the target group or listener.

\paragraph{Offensiveness:}
Offensiveness captures the degree of offensiveness or harm that a statement can cause to the target group or listener. The offensive dimension is represented as a free-text description of the type of harm or effect involved in the statement, such as “racism”, “sexism”, or “homophobia”.

\section{Instructions}
\label{sec:appendix-instructions}
\input{tables/instructions_table}
While defining Instructions for both the Auxiliary Explanation Generation and Counterspeech Generation tasks, we follow a standard format where the instruction is brief, to the point, and describes the expectation from the model clearly (See Table \ref{tab:instructions_table}). 

\section{Prompting}
\label{sec:appendix-prompting}
\input{tables/prompting_templates}
Drawing from previous studies \cite{lee2023rlaif}, we follow a \textit{preamble prefix} (\textbf{Preamble - Instruction - Exemplar}) prompt template for both the zero-shot and few-shot experiments. An example prompt template for one-shot exemplar is illustrated in Table \ref{tab:PROMPT_TEMPLATES}. For each of the three LLM baselines described in section \ref{sec:baselines}, we conduct inference on the IntentCONANv2 test set. For few-shot prompting, we sample exemplars for in-context learning from the IntentCONANv2 training set. For a given hate speech instance from the test set, we select exemplars based on the top-n semantically similar instances from the training set based on BM25 scores. We experiment with n=$3$, n=$5$, and n=$8$ exemplars, are find that the best results are obtained from n=3 in-context examples. We report the numbers with n=3 in-context examples for all few-shot experiments.

For inference on the FLAN-T5 XXL model, we make use of one NVIDIA A100 (80 GB) GPU, which results in $2$ inferences per second. For GPT-3.5-turbo, we use the OpenAI API\footnote{\url{https://openai.com/blog/openai-api}}. On average each turn takes about $2$ seconds for inference for both zero- and few-shot inference.

\section{Reward Model (RM)}
\label{sec:appendix-reward-model}
As detailed under section \ref{sec:phase3}, we design a composite reward function by combining outputs of three transformer-based models - Argument Quality (AQ), Pro-Con (PC), and Toxicity (T).

\paragraph{Argument Quality:} 
Akin to the Argument Quality (AQ) service detailed by \citet{bar-haim-etal-2021-project-debator}, we train a BERT-based regression model using a dataset comprising around $27,000$ arguments, which span a diverse range of subjects. Each argument in this dataset has been assigned a quality rating, as outlined in \cite{Gretz_Friedman_Cohen-Karlik_Toledo_Lahav_Aharonov_Slonim_2020}. High-quality arguments are characterized by their grammatical accuracy, proper use of language, clear and concise expression, and a clearly defined stance on the topic.

In this study, the primary role of the regression model is to assess the effectiveness of counterspeech. It evaluates how well a counterspeech can form logical and coherent arguments. The model assigns a score ranging from $0$ to $1$ to each argument, where a score of $1$ represents a high-quality argument and a score of 0 indicates a low-quality one. Essentially, this scoring system is used to determine the effectiveness of counterspeech in terms of its argumentative quality and coherence.

\paragraph{Pro-Con:} Similar to the Pro-Con service described by \citet{bar-haim-etal-2021-project-debator}, we train a BERT-based classifier on 400K stance labeled examples, including arguments extracted from the Lexis-Nexis corpus, as well as arguments collected via crowdsourcing \cite{toledo-ronen-etal-2020-multilingual}. The classifier's function is to evaluate a given sentence's stance towards a specific topic, assigning a score between $-1$ and $1$. A score of $1$ signifies a pro-stance (in favor of the topic), while $-1$ indicates a con-stance (contradicting the topic). In the context of this study, the classifier is used to analyze the relationship between hate speech (as the topic) and counterspeech (as the sentence). The goal is to determine how well the counterspeech opposes the hate speech. A high reward is expected if the counterspeech effectively contradicts the hate speech, and a lower reward if it supports it.

To facilitate this assessment, the output score from the Pro-Con classifier, denoted as $(PC(x', y'))$ (Refer section \ref{sec:appendix-reward-model}), is normalized from its original range of $[-1, 1]$ to $[0, 1]$ by $(\frac{1 - PC(x', y')}{2})$. This normalization ensures that lower Pro-Con scores, which indicate effective counterspeech against hate speech, correspond to higher rewards. This approach aligns the Pro-Con scoring mechanism with the reward structure desired in the study, emphasizing the efficacy of counterspeech in opposing hate speech.

\paragraph{Toxicity:} 
We use a pre-trained model from the Detoxify python library\footnote{\href{4}{https://pypi.org/project/detoxify/}}. Specifically, we employ the texttt{unbiased} classifier, which is a RoBERTa-based model trained on the Jigsaw Unintended Bias in Toxicity Classification challenge\footnote{\href{5}{https://www.kaggle.com/c/jigsaw-toxic-comment-classification-challenge}}. The primary function of this model is to predict the level of toxicity in a given text input. It generates a score ranging from $0$ to $1$, where $0$ corresponds to low toxicity and $1$ to high toxicity. 

In the composite reward function, the model's output is utilized to determine the toxicity of counterspeech that is generated. The objective is to ensure that counterspeech has minimal toxicity. Therefore, within the reward function of the study, the toxicity score is subtracted from $1$, effectively inverting the scale. In this inverted scale, a higher value is considered better, signifying lower toxicity.

\section{Experimental Setup}
All the models in our study were developed using the HuggingFace platform\footnote{\href{6}{https://huggingface.co/}} and Pytorch. For the implementation of LoRA, we utilized the \texttt{peft} Python library\footnote{\href{7}{https://pypi.org/project/peft/}}, and for the Proximal Policy Optimization (PPO) algorithm, the TRL library was employed\footnote{\href{8}{https://github.com/lvwerra/trl}}. At inference time, we use a common set of generation parameters in all our experiments: top\_k = $1$, top\_p = $1$, max\_new\_tokens = $512$, temperature = $1$, do\_sample = False. The entirety of our training and testing was conducted on a single NVIDIA A100 (80 GB) GPU.  Detailed descriptions of the training procedures for each phase of \texttt{CoARL} are presented in the subsequent sections.

\paragraph{Phase 1: Auxiliary Explanation Generation.}
In this phase, we fine-tuned a FLAN-T5 XXL model, which has $11$ billion parameters, over three epochs. The training used the AdamW optimizer, with a learning rate of $1e-4$ and a batch size of 8. Beam search was employed as the decoding strategy. The entire training process, using FP32 on an NVIDIA A100 GPU, took approximately $12$ hours.

\paragraph{Phase 2: Task-Specific Adapter Training.}
During this phase, the model weights established in Phase 1 were locked, and we initialized a LoRA adapter using PEFT. This adapter was then fine-tuned on the \texttt{IntentConanV2} training set, following the instruction format detailed in Table \ref{tab:instructions_table}. The fine-tuning process involved specific hyperparameters, including a batch size of $8$, training for $16$ epochs, and using the Adam optimizer. The learning rate was set to $4e-6$. Other vital settings included a maximum input token length of $256$, a LoRA (r) value of $16$, a LoRA ($\alpha$) value of $32$, and a LoRA layer dropout rate of $0.05$.

\paragraph{Phase 3: Optimization via RLAIF.}
As detailed in section \ref{sec:phase3}, we trained a policy model using the Proximal Policy Optimization (PPO) method. This training applied to FLAN-T5 XXL with LoRA adapters from the previous phase, involved the following hyperparameters: a learning rate of $1.4e-6$, an initial KL penalty coefficient ($Init_kl_coeff$) of $0.03$, a batch size of $32$, a mini-batch size of $2$, $15000$ steps, a target KL value (Target) of $5$ for adaptive KL control, a horizon of $10000$ for adaptive KL control, a cliprange value of $0.25$ for loss calculation, $5$ epochs, and a target KL ($Target_kl$) of $0.05$.

%% file: figures/dataset_analysis_appendix.tex
\begin{figure}[t]
  \centering

  \begin{subfigure}{.5\linewidth}
    \centering
    \includegraphics[width=\linewidth]{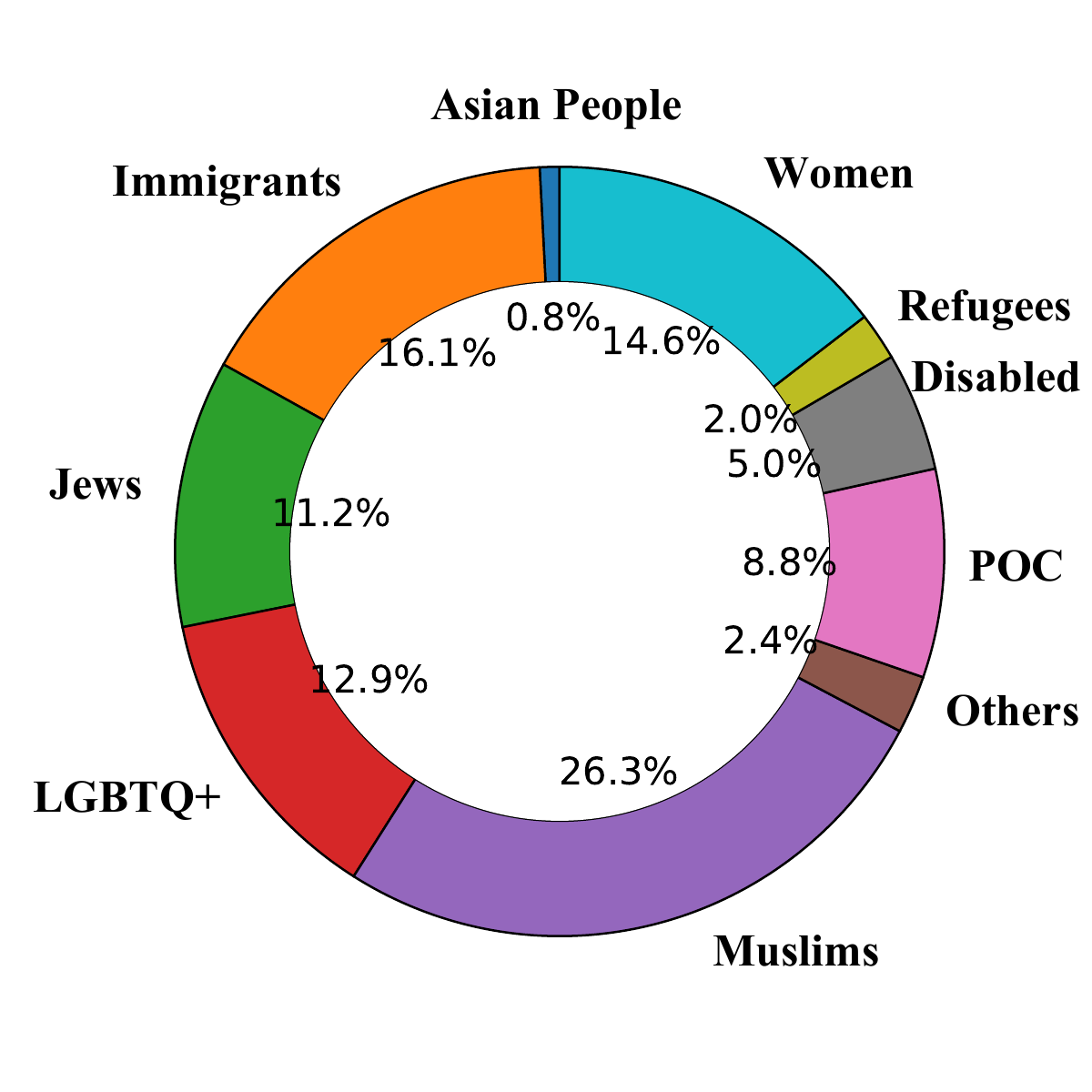}
    \caption{HS Target distribution}
    \label{4_1}
  \end{subfigure}%
  \begin{subfigure}{.5\linewidth}
    \centering
    \includegraphics[width=\linewidth]{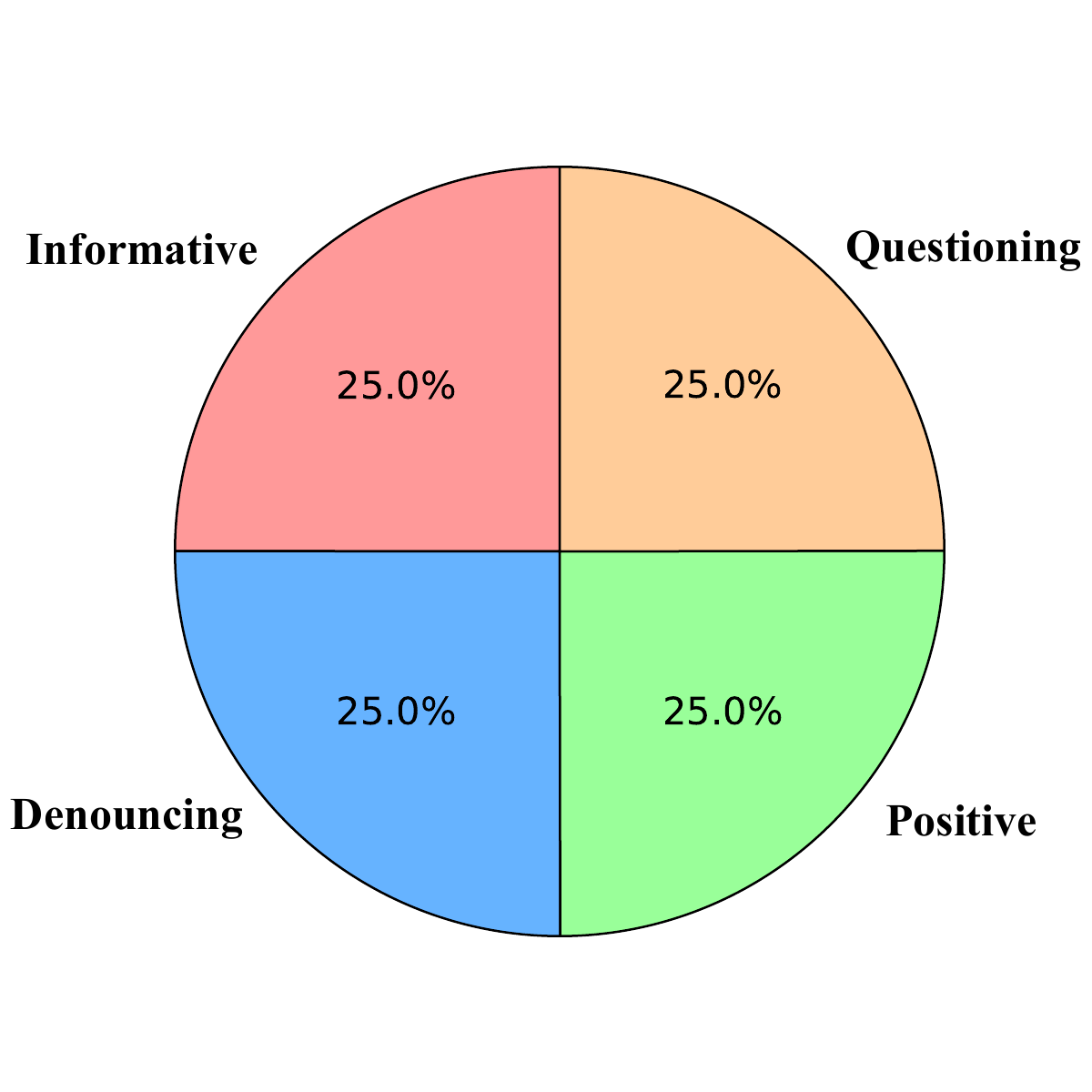}
    \caption{CS Intent distribution}
    \label{4_2}
  \end{subfigure}
  
  \medskip
  
  \begin{subfigure}{.5\linewidth}
    \centering
    \includegraphics[width=\linewidth]{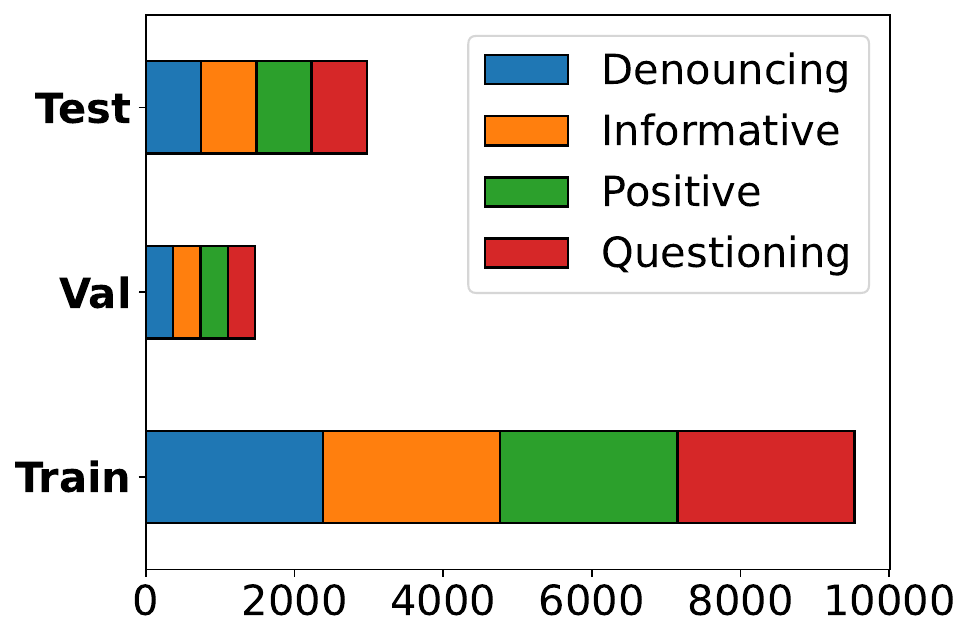}
    \caption{CS Intent distribution}
    \label{4_3}
  \end{subfigure}%
  \begin{subfigure}{.5\linewidth}
    \centering
    \includegraphics[width=\linewidth]{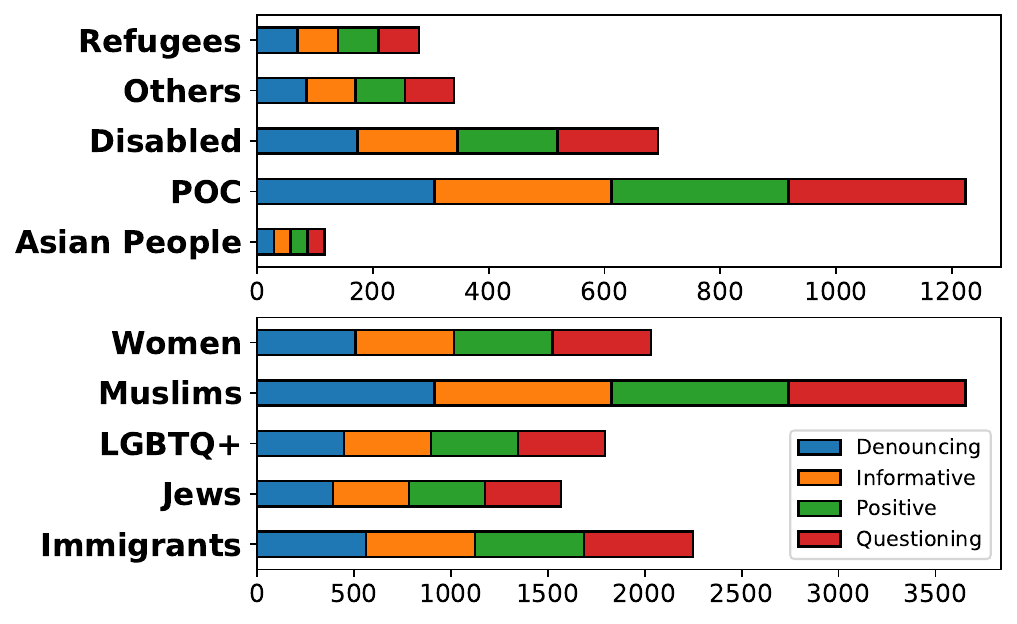}
    \caption{Intent vs Target}
    \label{4_4}
  \end{subfigure}
  
  \medskip
  
  \begin{subfigure}{.35\linewidth}
    \centering
    \includegraphics[width=\linewidth]{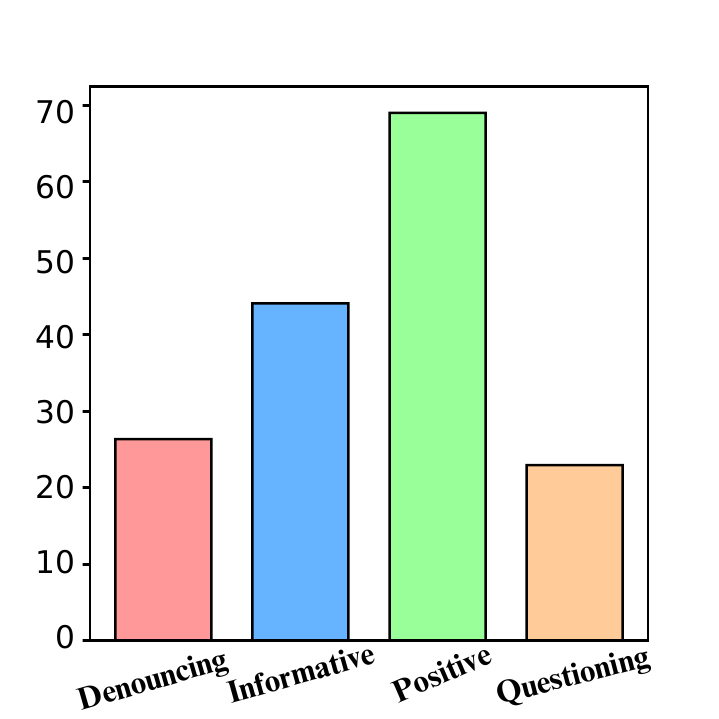}
    \caption{Intents: Mean token length}
    \label{4_5}
  \end{subfigure}%
  \begin{subfigure}{.73\linewidth}
    \centering
    \includegraphics[width=\linewidth]{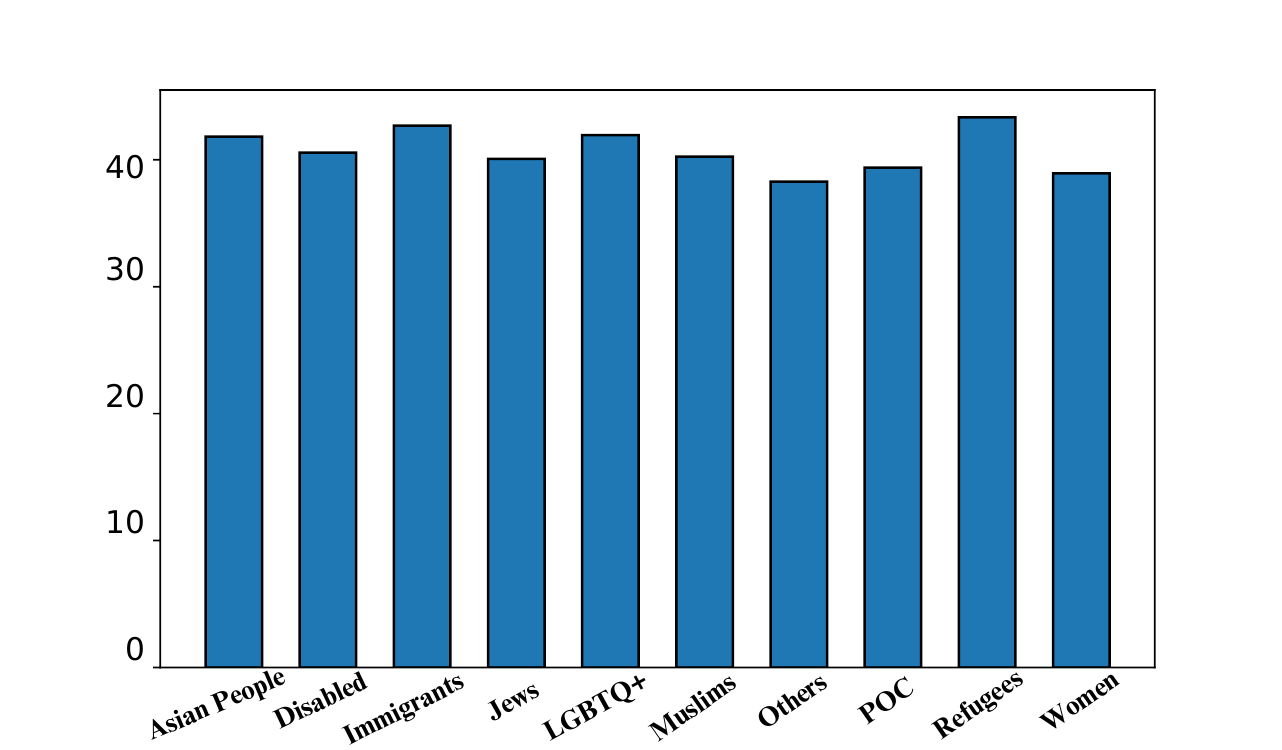}
    \caption{Targets: Mean token length}
    \label{4_6}
  \end{subfigure}
  
  \caption{Visual exploration of various attribute distribution present in \texttt{IntentCONANv2}}
  \label{4_all}
\end{figure}

%% file: tables/dataset_examples.tex
\begin{table}[t!]
\centering
\scalebox{0.78}{%
\begin{tabular}{@{}lll@{}}
\toprule
\textbf{HS}                      &                                   & Gypsies are essentially thieves.                                                                                                                                                                                                                                                                                                                                                                                                                                                    \\ \midrule
\multicolumn{1}{l|}{}            & \multicolumn{1}{l|}{\textbf{INF}} & \begin{tabular}[c]{@{}l@{}}Stereotyping an entire community based on\\ the actions of a few is unfair and perpetuates\\ discrimination. it is important to recognize\\ that criminal behavior exists in every society\\ and is not limited to any specific group.\\ Painting all gypsies as thieves is a harmful\\ generalization. It is crucial to approach\\ individuals and communities with respect\\ and empathy rather than perpetuating\\ negative stereotypes.\end{tabular} \\ \cmidrule(l){2-3} 
\multicolumn{1}{l|}{\textbf{CS}} & \multicolumn{1}{l|}{$\textbf{QUE}^*$} & \begin{tabular}[c]{@{}l@{}}They are not thieves, just like any other citizen.\\ Did you just infer that from their belonging to\\ that community?\end{tabular}                                                                                                                                                                                                                                                                                                                      \\ \cmidrule(l){2-3} 
\multicolumn{1}{l|}{}            & \multicolumn{1}{l|}{\textbf{DEN}} & \begin{tabular}[c]{@{}l@{}}Respect and understanding should prevail\\ instead.\end{tabular}                                                                                                                                                                                                                                                                                                                                                                                         \\ \cmidrule(l){2-3} 
\multicolumn{1}{l|}{}            & \multicolumn{1}{l|}{\textbf{POS}} & \begin{tabular}[c]{@{}l@{}}It is important to avoid generalizations and\\ stereotypes about any community, including\\ the gypsy community. Stereotyping an entire\\ group based on the actions of a few individuals\\ is unfair and perpetuates discrimination. Let’s\\ focus on promoting understanding and empathy\\ towards diverse communities, appreciating their\\ rich cultural heritage and contributions.\end{tabular}                                                    \\ \bottomrule
\end{tabular}%
}
\caption{Comparison between IntentCONAN and IntentCONANv2. The $*$ shows the pre-existing counter speech in the IntentCONAN dataset.}
\label{comparison table}
\end{table}

%% file: tables/instructions_table.tex
\begin{table*}[t]
  \centering
  \begin{tabular}
  {lp{3cm}p{10cm}} 
    \toprule
    Task-ID & Generation Type & Instruction \\
    \midrule
    I$_1$ & Offensiveness & Analyze the offensiveness of the statement: \{\texttt{HS}\} \\
    \noalign{\vskip 1.0ex} 
    I$_2$ & Target Group & Identify the group of people that the speaker is targeting or discriminating against in the offensive statement: \{\texttt{HS}\} \\
    \noalign{\vskip 1.0ex} 
    I$_3$ & Speaker Intent & Analyze the speaker's intention behind writing the offensive statement: \{\texttt{HS}\} \\
    \noalign{\vskip 1.0ex} 
    I$_4$ & Power Dynamics & Explain the underlying power dynamics between the speaker and the target group in the offensive statement: \{\texttt{HS}\} \\
    \noalign{\vskip 1.0ex} 
    I$_5$ & Implication & Explain the implied meaning underlying the offensive statement: \{\texttt{HS}\} \\
    \noalign{\vskip 1.0ex} 
    I$_6$ & Emotional Reaction & Describe how the target group might feel emotionally after reading or listening to the offensive statement: \{\texttt{HS}\} \\
    \noalign{\vskip 1.0ex} 
    I$_7$ & Cognitive Reaction & Describe how the target group might react cognitively after reading or listening to the offensive statement: \{\texttt{HS}\} \\
    \midrule
    \noalign{\vskip 0.5ex}
    I$_8$ & Intent-Specific Counterspeech & Analyze the different aspects such as offensiveness, target group, stereotype, power dynamics, implied meaning, emotional, and cognitive reactions before writing a \{\texttt{INT}\} counterspeech for the offensive statement: \{\texttt{HS}\} \\
    \bottomrule
  \end{tabular}
\caption{Detailed Instructions for the tasks of explanation and counterspeech generation respectively. The instructions labeled \{I$_1$, I$_2$, ..., I$_7$\} correspond with each of the dimensions of hate speech explanations in the Auxiliary Explanation Generation (AEG) task, as outlined in Section \ref{sec:phase1}. Instruction I$_8$ is crafted for training task-specific LoRA adapter for generating intent-conditioned counterspeech, detailed in Section \ref{sec:phase2}. In these instructions, {\texttt{HS}} represents the instance of hate speech, and {\texttt{INT}} denotes the targeted intent of the counterspeech, which includes \texttt{Positive}, \texttt{Denouncing}, \texttt{Informative}, or 
\texttt{Questioning}.}

\label{tab:instructions_table} 
\end{table*}

%% file: tables/prompting_templates.tex
\begin{table*}[!ht]
\begin{center}
\small 
\setlength{\tabcolsep}{2pt} 
\renewcommand{\arraystretch}{1.0} 
\resizebox{\textwidth}{!}{%
\begin{tabular}{p{1.3in}p{5.2in}}
\toprule
\textbf{\texttt{Preamble}} &
  \begin{tabular}[c]{@{}p{5.2in}@{}}
  \texttt{A Positive Counterspeech responds to a hate speech statement using empathy and affiliation…attenuating entrenched extremist viewpoints.}\\\\
  \texttt{A Counterspeech is considered to be good if it satisfies certain desired qualities like relevance, effectiveness, and non-toxicity. Below we define each of these axes:}\\\\
  \texttt{\textbf{Relevance:} This axis evaluates the extent to which counterspeech directly addresses the core message of the hate speech. It assesses whether the counterspeech maintains a coherent and focused dialogue aimed at neutralizing the impact of the hate speech. A low relevance indicates a disconnect between the counterspeech and the hate speech, exemplified by counterspeech that strays off-topic. For instance, if the hate speech centers on LGBTQ issues, but the counterspeech diverts to religious beliefs, it would be deemed irrelevant.}\\
  \texttt{\textbf{Effectiveness:} This axis gauges the ability of counterspeech to logically and coherently challenge or refute the biases, stereotypes, or offensive content present in the hate speech. Effective counterspeech should logically counteract the hate speech, thereby diminishing its impact or negating its offensive message.}\\
  \texttt{\textbf{Non-toxicity:} This axis measures the level of respectfulness and reasonableness in the counterspeech. It is essential that counterspeech remains free from rudeness or provocativeness. A non-toxic approach is crucial as counterspeech that is perceived as aggressive or disrespectful may exacerbate the situation, potentially leading to an escalation of hate speech rather than its mitigation.}
  \end{tabular} \\
\midrule
\textbf{\texttt{Instruction}} &
  \begin{tabular}[c]{@{}p{5.2in}@{}}
  \texttt{»»»» Instruction »»»»}\\
  \texttt{Given a hate speech statement, generate a \textbf{Positive Counterspeech} by following the definitions given above.}
  \end{tabular} \\
\midrule
\textbf{\texttt{Exemplars (1-shot)}} &
  \begin{tabular}[c]{@{}p{5.2in}@{}}
  \texttt{»»»» Examples »»»»}\\
  \texttt{\textbf{Statement} – Children should learn about gender identity and sexual orientation from their families rather than in class from LGBT fundamentalists.}\\
  \texttt{\textbf{Positive Counterspeech} – Teaching all children to be accepting of each other gives a sense of belonging to any child who feels a little bit different, and helps them to know it’s okay to be themselves.}
  \end{tabular} \\
\midrule
\textbf{\texttt{Sample to Generate}} &
  \begin{tabular}[c]{@{}p{5.2in}@{}}
  \texttt{\textbf{Statement} – LGBTs just want to brainwash our children. They should not be taught about gender identity and sexual orientation in school, theyll end up being bullied.}\\
  \texttt{\textbf{Positive Counterspeech} –}
  \end{tabular} \\
\bottomrule
\end{tabular}%
}
\end{center}
\caption{Prompt template used for zero-shot and few-shot experiments. }
\label{tab:PROMPT_TEMPLATES}
\end{table*}

%% file: main.bbl
\begin{thebibliography}{62}
\bibitem[{Leetaru(2019)}]{leetaru-2019}
Kalev Leetaru. 2019.
\newblock Online toxicity is as old as the web itself but the return to communities may help. forbes magazin.
\end{thebibliography}
